\documentclass[11pt,a4paper]{article}
\usepackage[hyperref]{emnlp2020}
\usepackage{times}
\usepackage{latexsym}

\usepackage{amssymb,amsmath}
\usepackage{graphicx}
\usepackage{subcaption}
\usepackage{tikz}
\usepackage{pgfplots}
\usepackage{hyperref}
\usepackage{enumitem}

\usepackage{url}
\usepackage{multirow}

\usepackage{microtype}

\aclfinalcopy 

\title{Predicting Event Time by Classifying Sub-Level Temporal Relations Induced from a Unified Representation of Time Anchors}

\author{Fei Cheng \\
  Department of Intelligence \\
  Science and Technology \\
  Graduate School of Informatics \\
  Kyoto University \\
  \texttt{feicheng@i.kyoto-u.ac.jp} \\\And
  Yusuke Miyao \\
  Department of Computer Science \\
  Graduate School of Information \\
  Science and Technology \\
  University of Tokyo \\
  \texttt{yusuke@is.s.u-tokyo.ac.jp} \\}
  
\date{}

\begin{document}
\maketitle
\begin{abstract}
Extracting event time from news articles is a challenging but attractive task. In contrast to the most existing pair-wised temporal link annotation, \citet{reimers-dehghani-gurevych:2016:P16-1} proposed to annotate the time anchor (a.k.a. the exact time) of each event. Their work represents time anchors with discrete representations of {\it Single-Day/Multi-Day} and {\it Certain/Uncertain}. This increases the complexity of modeling the temporal relations between two time anchors, which cannot be categorized into the relations of Allen's interval algebra~\citep{allen1990maintaining}. 

In this paper, we propose an effective method to decompose such complex temporal relations into sub-level relations by introducing a unified quadruple representation for both \emph{Single-Day/Multi-Day} and \emph{Certain/Uncertain} time anchors. The temporal relation classifiers are trained in a multi-label classification manner. The system structure of our approach is much simpler than the existing decision tree model~\cite{reimers2018event}, which is composed by a dozen of node classifiers. Another contribution of this work is to construct a larger event time corpus (256 news documents) with a reasonable Inter-Annotator Agreement (IAA), for the purpose of overcoming the data shortage of the existing event time corpus (36 news documents). The empirical results show our approach outperforms the state-of-the-art decision tree model and the increase of data size obtained a significant improvement of performance.

\end{abstract}

\section{Introduction}

Along with TimeBank~\citep{pustejovsky2003timebank} (TB) and other temporal corpora, a series of competitions on temporal information extraction (TempEval-1,2,3)~\citep{verhagen2009tempeval,verhagen2010semeval,uzzaman2012tempeval} are attracting growing research efforts. Predicting the exact time when an event occurred can effectively contribute various time-aware tasks, such as timeline~\citep{minard-EtAl:2015:SemEval}, temporal question answering~\citep{llorens-EtAl:2015:SemEval,meng-rumshisky-romanov:2017:EMNLP2017}, knowledge base population, etc. For instance, given a hot topic (event, person, product, company, etc.), we can first gather its related news articles, extract events, and predict event time from them. An event timeline can be automatically generated by ordering events on a time axis according to their occurring time. Such timeline representations can be extremely helpful for readers to comprehend new topics efficiently.

TimeBank-EventTime (TBET) is a new temporal information corpus based on TB. Instead of using the majority of a pair-wise Temporal Link (TLINK)~\cite{setzer2002temporal} to model a relation between two mentions\footnote{In this paper, 'Timex' denotes time expression and 'DCT' denotes document creation time} (i.e. event, time expression and document creation time), TBET annotates the occurring time of each individual event. Their annotation schema first distinguishes between \emph{Single-Day/Multi-Day} and \emph{Certain/Uncertain} events as following:

\begin{itemize}
\item \textbf{\emph{Single-Day}} denotes an event occurs in a single day. A \emph{Single-day} event can be further categorized into two types as follows:

\textbf{\emph{Certain}} day, e.g. \textit{'1998-02-06'}

\textbf{\emph{Uncertain}} day, e.g. \textit{'before1998-01-31'}, \textit{'after1998-01-01'} or \textit{'after1998-01-01, before1998-02-06'}.

\item \textbf{\emph{Multi-Day}} event time can be seen as a tuple of two \emph{Single-Day} \textit{(begin, end)}.
\end{itemize}

For predicting \emph{Single-Day} event time, \citet{N18-1166} induced 6-label temporal relations induced from TBET to train two classifiers of Event-to-DCT and Event-to-Timex mention pairs. A major difficulty preventing them from extending this approach to the \emph{Multi-Day} events is the complexity of inducing relations with \textit{Uncertain} information. Manually extending the existing temporal relation types (\citet{allen1990maintaining}'s interval algebra) will be hard.

Another approach~\citep{reimers2018event} of predicting both \emph{Single-Day} and \emph{Multi-Day} event time trains a series of classifiers to make tree-structure decisions starting from the root decision of whether an event is \emph{Single-Day} or \emph{Multi-Day}. The \emph{Single-Day} branch determines the temporal clues from relevant time expressions. While, the \emph{Multi-Day} branch will further determine begin and end points separately. However, such decision tree of classifiers is less generalized for only fitting this event time definition and training a series of classifiers is complex and time-consuming.

In this work, we propose a simple method to predict both \emph{Single-Day} and \emph{Multi-Day} event time effectively. We introduce a quadruple expression with 4-date values, i.e. \textit{(begin\_earliest}, \textit{begin\_latest}, \textit{end\_earliest} and \textit{end\_latest)}, which is capable of unifying the representation of \emph{Single/Multi-Day} and \emph{Certain/Uncertain} event time (Section~\ref{subsec:qrep}). This expression enables us to decompose a complex temporal relation into four simple sub-level temporal relations (Section~\ref{subsec:sr}). In Section~\ref{sec:system}, we perform classifications in a multi-label fashion. In the inference step, we gather all the relevant time clues of an event to infer the time. Another contribution (Section~\ref{sec:corpus}) is that we constructed a larger event time corpus with 256 news articles, compared to the existing TBET with 36 articles. In Section~\ref{subsec:dataincrease}, we show the experiments of increasing the training data and being evaluated in two different test data. The results show that the large data size significantly improves classification results in the large test data: TE3-Test. The comparison to the SOTA decision tree model shows our sub-level temporal relation model with attention significantly outperforms their model in the TBET data (Section~\ref{subsec:sota}).

\section{Background}

\subsection{Event Time Annotation}
\label{subsec:bg:eta}

Most existing temporal corpora adopted a pair-wise schema (i.g. TLINK) to encode temporal relations between mentions. A major drawback is that the potential pair candidates are quadratic to the number of mentions. This resulted in the sparsity of TLINKs in the original TimeBank. The later TimeBank-Dense~\citep{cassidy2014} achieved denser TLINKs by forcing  the annotators labeling all the possible pairs in any two adjacent sentences. However, the dense annotation is time-consuming and performed on a subset of Timebank with only 36 documments.

\citet{reimers-dehghani-gurevych:2016:P16-1} proposed an event time schema to annotate the TBET corpus with a linear annotation effort, in which annotators infer the exact time of each individual event. The main limitation is that the new schema requires events to be capable of being anchored on exact time. In this paper our main target is processing news articles, in which time expressions and document creation time provide rich exact time information. Therefore, this limitation will affect less.

\subsection{Existing Event Time Prediction Systems}

The proposal of TBET raised a new task of predicting the exact time of events. Intuitively, knowing when an event occurred is an attractive target to achieve many practical applications. For instance, ordering events according to their occurring time can naturally generate an event timeline in a cross-document scenario. 

\citet{reimers-dehghani-gurevych:2016:P16-1} released a baseline for event time prediction based on CAEVO~\cite{chambers2014dense}, which is trained on TimeBank-Dense. This baseline can only predict \emph{Single-Day} Events, because the Timebank-Dense TLINKs adopted a limited 6-label set without providing the necessary begin or end information to predict \emph{Multi-Day} Events.

\citet{N18-1166} proposed an approach to induce  similar 6-label relations based on the event time annotation. Their temporal relation classifiers are fully trained on the new TBET corpus and achieve better performance on predicting \emph{Single-Day} event time. 

\citet{reimers2018event} included the \emph{Multi-Day} Events prediction into their target. They designed a series of individual classifiers to make hierarchical decisions of whether an event is \emph{Single-Day} or \emph{Multi-Day}, the temporal relation from the begin day of an event to a time expression, etc. The drawback is that training a decision tree of individual classifiers is complicated and the data of some leaf classifiers become very sparse.

\begin{table*}[t]
  \begin{center}
    \begin{tabular}{l|c|c|c} 
      \textbf{\emph{Single/Multi}} & \textbf{\emph{Certainty}} & \textbf{Type} & \textbf{Representation}  \\
      \hline
      \multirow{2}{*}{\emph{Single-Day}} & \multirow{2}{*}{\emph{Certain}} & \emph{Time anchor} & 1998-01-26 \\
       & & \emph{Quadruple} & ((1998-01-26, 1998-01-26), (1998-01-26, 1998-01-26)) \\ \hline
       \multirow{2}{*}{\emph{Single-Day}} & \multirow{2}{*}{\emph{Uncertain}} & \emph{Time anchor} & after1998-01-26before1998-02-06 \\
       & & \emph{Quadruple} & ((1998-01-26, 1998-02-06), (1998-01-26, 1998-02-06)) \\ \hline
       \multirow{2}{*}{\emph{Multi-Day}} & \multirow{2}{*}{\emph{Certain}} & \emph{Time anchor} & begin:1998-01-01,end:1998-01-31 \\
       & & \emph{Quadruple} & ((1998-01-01, 1998-01-01), (1998-01-31, 1998-01-31)) \\ \hline
       \multirow{2}{*}{\emph{Multi-Day}} & \multirow{2}{*}{\emph{Uncertain}} & \emph{Time anchor} & begin:1998-01-01,end:after1998-02-06 \\
       & & \emph{Quadruple} & ((1998-01-01, 1998-01-01), (1998-02-06, \detokenize{~})) \\ \hline
    \end{tabular}
    \caption{\label{tab:rep} The comparison of the time anchor and quadruple examples. \detokenize{'~'} denotes the blank information in some \emph{Uncertain} cases.}
  \end{center}
\end{table*}

\subsection{Neural Temporal Relation Classifier}

Temporal relation classification can be categorized as a variation of a more general Relation Classification (RC) task, in which the models are required to detect a semantic relation between two nominals in a sentence. \citet{C14-1220} proposed a Convolutional Neural Network (CNN) model with the relevant distances from each word to the two nominals as the additional features to achieve state-of-the-art performance. From 2016, researchers started to introduce the attention mechanism into RC to push the scores higher. \citet{zhou-etal-2016-attention} proposed an attention-based RNN model to obtain normalized scores along the time steps for calculating a weighted representation of an sentence. 

The step of applying neural networks into Temporal Relation Classification started from the work of~\citep{cheng-miyao:2017:Short,meng-rumshisky-romanov:2017:EMNLP2017}. These two works feed the shortest dependency path (SDP) between two mentions into recurrent neural networks (RNNs). \citet{reimers2018event} adopted \citet{C14-1220}'s model to train a series classifiers to make hierarchical decisions. 

In this work, we proposed a mention-wised attention LSTM~\cite{hochreiter1997long} classifier, which emphasized the interaction between words and the given mentions to obtain the weighted context information of an sentence.

\section{Representing Time Anchors and Sub-level Temporal Relations}
\label{sec:def}

There are two obstacles to prevent us from achieving a more effective way to predict event time. First, the discrete annotation of the \emph{Single/Multi-day} and \emph{Certain/Uncertain} information in TBET forced \citet{reimers2018event} to make hierarchical decisions from the top question whether the given event is \emph{Single-Day} or \emph{Multi-Day} to the leaf temporal relation classifiers. It raised the complexity of their system and accumulated errors from early classifiers. The second question is that extending the definition of temporal relation is hard but necessary in order to represent the relation with \emph{Uncertain} time.

\subsection{A Unified Quadruple Representation of Time Anchors}
\label{subsec:qrep}

In this work, we adopt a quadruple representation \cite{berberich2010language} for consistently representing the \emph{Single/Multi-Day} and \emph{Uncertainty} information of a time anchor.

\[ T = ((begin_e, begin_l), (end_e, end_l)) \]

$(begin_e, begin_l)$ stand for the earliest and latest possible days of the beginning point. Analogously, $(end_e, end_l)$ are its earliest and latest possible days of the end point. 

This representation is naturally designed for representing \emph{Multi-Day} and \emph{Uncertain} information. We adopt an inclusive rule, which includes \emph{Single-Day} into our representation by making beginning and end annotation to the same. A \emph{Certain} day has the same earliest and latest days. To make it clear, we list several examples in Table~\ref{tab:rep} with both TBET and our quadruple representations for comparison.  

\begin{figure*}[t]
\center{\includegraphics[width=0.98\linewidth]{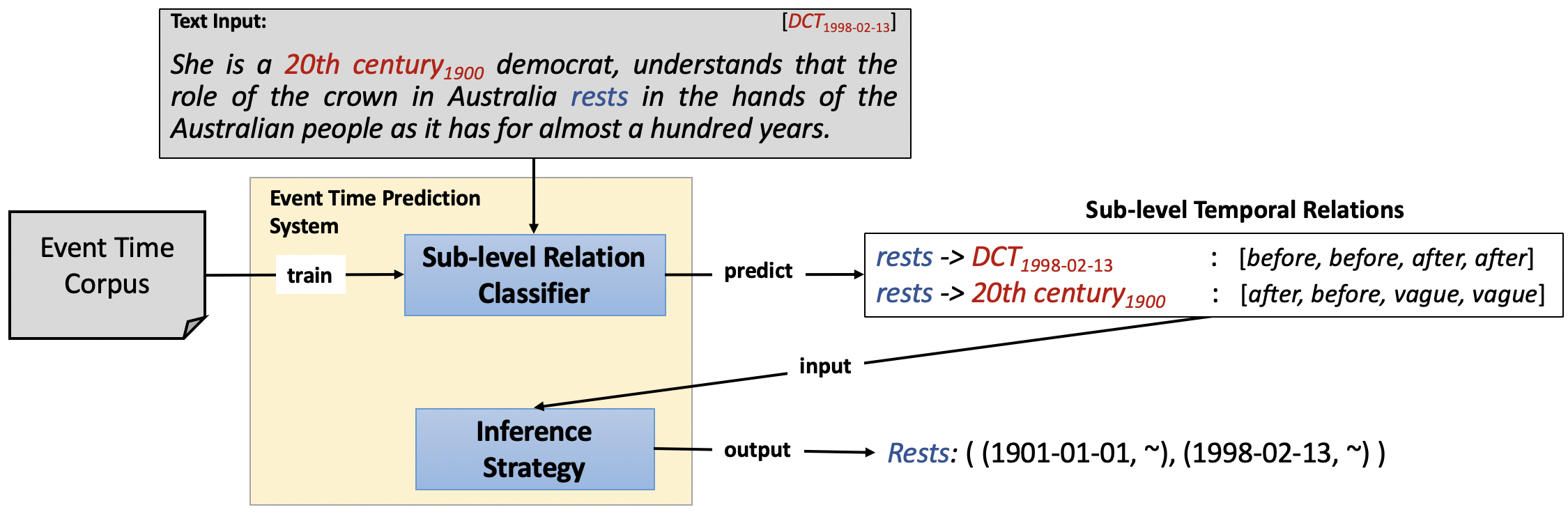}}
\caption{\label{fig:system} The architecture of the event time prediction System. }
\end{figure*}

\begin{figure}[!t]
\vspace{12pt}
\center{\includegraphics[width=0.98\linewidth]{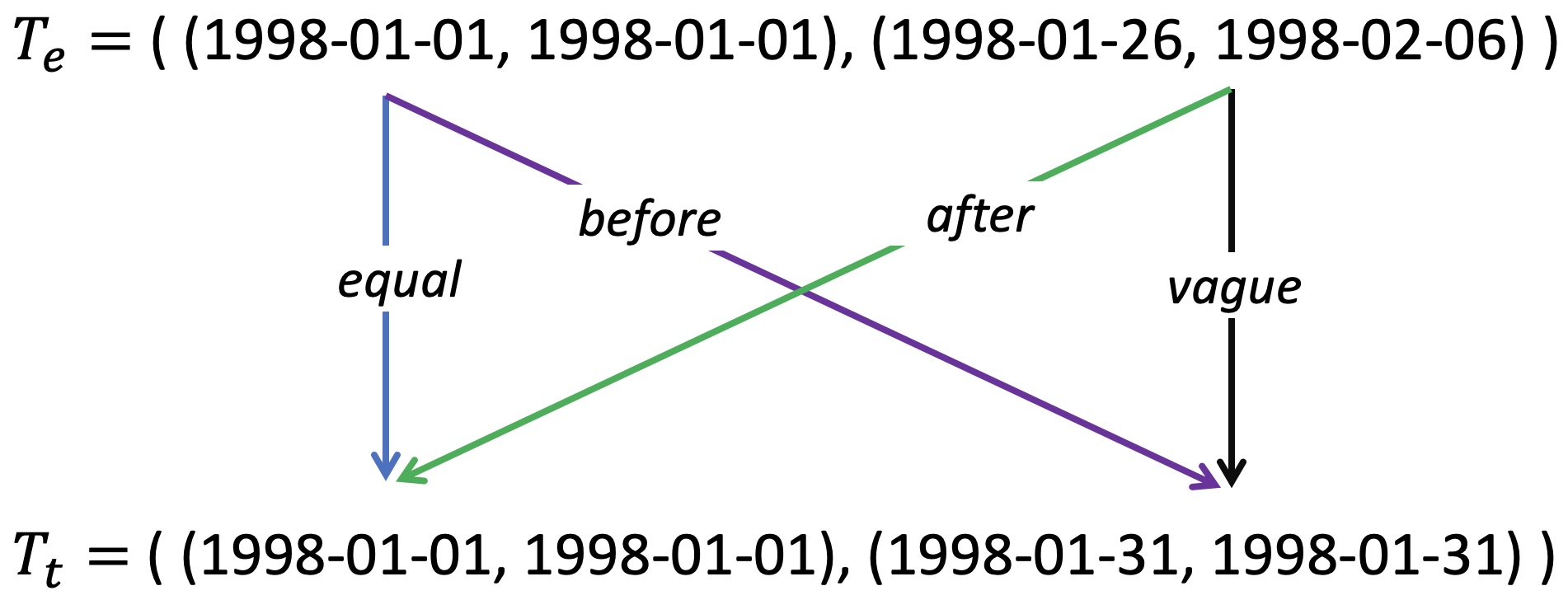}}
\caption{\label{fig:sublabel}An example of sub-level temporal relations between $T_e$ and $T_t$ }
\end{figure}

\subsection{Decomposing temporal relations into sub-level}
\label{subsec:sr}

Most existing temporal relation annotation adopted \citet{allen1990maintaining}'s algebra, which defined 13 relations between two \emph{Certain} time intervals. Many of the following works attempted to reduce the sparsity of relations by merging some relations to a coarse relation. The \emph{Uncertainty} information in TBET even aggravates the complexity of modeling temporal relations between two time anchors. 

In this work, we decompose a temporal relation into a group of sub-level relations for avoiding laboriously extending the relation definition. Let $((begin^e_e, begin^e_l), (end^e_e, end^e_l))$ and $((begin^t_e, begin^t_l), (end^t_e, end^t_l))$ stand for two time anchors: $T_e$ of an event and $T_t$ of a DCT or Timex. We explicitly compare four sub-level relations (SR) $[SR_1, SR_2, SR_3, SR_4]$ as follows:
\begin{itemize}[leftmargin=.54in]
	\item[$SR_1$:] $(begin^e_e, begin^e_l)$ to $(begin^t_e, begin^t_l)$
	\item[$SR_2$:] $(begin^e_e, begin^e_l)$ to $(end^t_e, end^t_l)$
	\item[$SR_3$:] $(end^e_e, end^e_l)$ to $(begin^t_e, begin^t_l)$
	\item[$SR_4$:] $(end^e_e, end^e_l)$ to $(end^t_e, end^t_l)$
\end{itemize}

Figure~\ref{fig:sublabel} is an example of the sub-level relations between $T_e$ and $T_t$. This example can not be categorized into any existing relation types defined in Allen's algebra due to the \emph{Uncertainty} of whether the end point of $T_e$ is {\it after}, {\it before} or {\it equal} to the end point of $T_t$. However, our proposal can encode the sub-level information $[equal, before, after, vague]$ indicating that $T_e$ and $T_t$ holds the same begin points, but the temporal relation between the end points is {\it vague}.

Each sub-level relation is naturally defined with a simple label set \{{\it equal}, {\it after}, {\it before}, {\it vague}\} as shown in Table~\ref{table:SRdef}. Any relation between two time anchors can be represented as the combination of four sub-level relations. This approach effectively avoids an exhaustive expansion of Allen's algebra in order to include the \emph{Uncertain} information of time anchors. 

\begin{table}[t!]
\centering
\small{
\begin{tabular}{@{}p{1.6cm}|p{5.6cm}@{}}\hline
\textbf{SR Label} & \textbf{Definition}\\ \hline \hline
\multicolumn{2}{l}{between $(E_e, E_l)$ and  $(T_e, T_l)$}  \\ \hline
{\it equal} & if $E_e = T_e$ and  $E_l = T_l$  \\
{\it after} & if $E_e \geq T_l$ and ($E_l > T_l$ or $E_l=\detokenize{~}$)  \\
{\it before} & if $E_l \leq T_e$ and ($E_e < T_e$ or $E_e=\detokenize{~}$)  \\

{\it vague} & other cases   \\ \hline 
\end{tabular}
}
\caption{Definition of four types of Sub-level temporal relations between $(E_e, E_l)$ and  $(T_e, T_l)$. `$<$', `$>$', `$=$' denote one day point is in the left of, right of and same position as the other day point in a left-to-right time axis. \label{table:SRdef}
}
\end{table}

\section{Event Time Prediction System}
\label{sec:system}

Once we have the unified representation of time anchors and sub-level relations (SR), we can directly classify SRs between a given event to DCT or any other Timex in the text, instead of constructing a complex tree structure of node classifiers. The overall system architecture is shown in Figure~\ref{fig:system}. Given a target event, our SR classifiers first predict four SRs to each time expression in the context. Then, the quadruple values of the given event are inferred by a strategy based on the SR inputs. 

\begin{figure}[t]
\center{\includegraphics[width=0.96\linewidth]{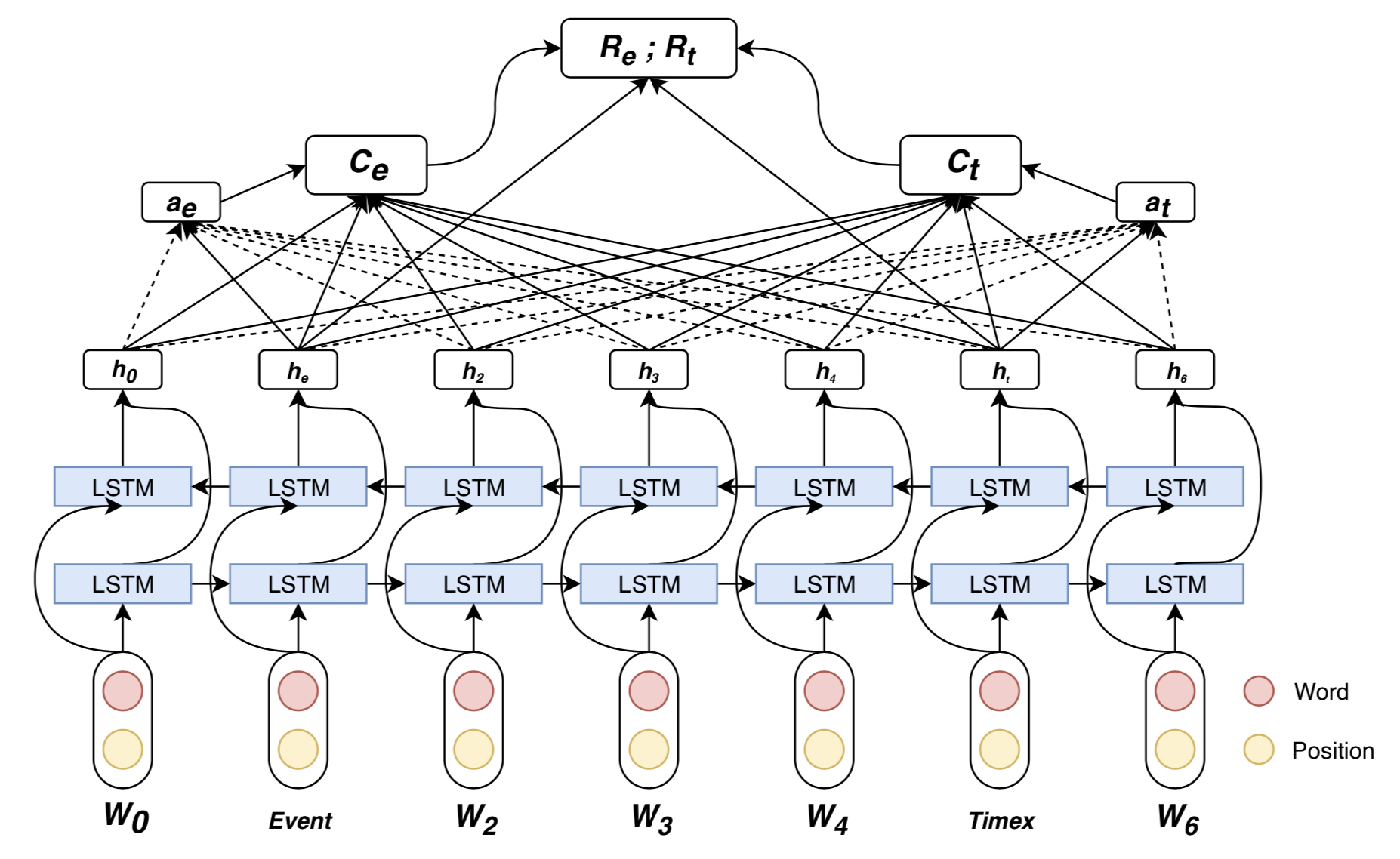}}
\caption{\label{fig:classifier} The mention-wised attention LSTM model. }
\end{figure}

\subsection{Mention-wise Attention Classifiers}

Based on the link types (Event-to-DCT and Event-to-Timex), temporal relation classifiers take different input information. For instance, Event-to-DCT (E-D) has only one event information, while Event-to-Timex (E-T) has two intputs, i.e. an event and a Timex. Therefore, we follow the setting of most existing work, which treats Event-to-DCT and Event-to-Timex as two independent classification tasks.

In this work, we propose a mention-wise attention LSTM classifier. An alignment vector $a_e$, whose length is the same as the number of time steps of the input sequence $s$, is derived by the interaction between each hidden state $\overline{h}_s$ and the hidden state $h_e$ of the given event.
\begin{equation} \label{eq_softmax}
a_e(s)=\frac{exp(score(h_e, \overline{h}_s))} { \sum_{s'}^{} exp(score(h_e, \overline{h}_{s'}))}
\end{equation}

We adopted a similar 'general' score function to \citep{luong-etal-2015-effective} as Eq~(\ref{eq_score}). $W_a$ denotes a matrix of trainable model parameters.
\begin{equation} \label{eq_score}
score(h_e, \overline{h}_s) = h^\mathsf{T}_e W_a \overline{h}_s
\end{equation}

Given $a_e$ as weights, a context vector $c_e$ is computed as the weighted average over the hidden states of all time steps. Then, we define the representation $R_m$ of a sequence $s$ given an event $e$ is the concatenation of $h_e$ and $c_e$, which is fed into the final Softmax layer. 

Figure~\ref{fig:classifier} shows the structure of our neural network classifier for Event-to-Timex. 
An Event-to-DCT relation has only an event mention information. Therefore, the model feed $R_e$ into the Softmax layer. In the case of Event-to-Timex, two mentions exist i.e. an event and a Timex. The model concatenate $R_e$ and $R_t$ as the output. 

\subsection{Word and Position Embeddings}

We use the pre-trained word embedding Glove\footnote{\url{https://nlp.stanford.edu/projects/glove/}} presented by \citet{pennington2014glove}. The input words are mapped into their embedding vectors by the embedding layer of our model. We use the 200-dimension Glove in our model.

In TRC, the given information of a data sample includes a sentence and mentions (event or Timex) inside it. An important feature is to distinguish between the mention words and other words inside the sentence. A popular method is called position embedding, which represents the relative distance from a word to the mention word. For instance, an event the $k$-th word (from left to right) in a sentence,  a word $i$ has a relative distance $i-k$ to the event. We embed this number into a random initialized vector and the dimension of the vector is a hyper-parameter.

\subsection{Multi-label Loss}

Based on our definition in Section~\ref{subsec:sr}, each E-D or E-T link consists of 4 SRs and each SR adopts a 4-label set \{$equal$, $after$, $before$, $vague$\}. In this paper, we formulate the SR classification as a multi-label multi-class problem. We fixed the dimension number of the output layer to $16$ ($4$ SRs $\times$ $4$ Labels) and calculate the Softmax distribution for each SR. We define the multi-label loss as the sum of the negative log-likelihood of each SR.

\begin{equation} \label{eq_loss}
L=- {\sum_{i} {\sum_{j} g_{ij} {\log p_{ij}}}}
\end{equation}

In Eq~\ref{eq_loss}, $i$, $j$ denote the SR, label. $g$, $p$ denote the gold label, predicted Softmax probability.

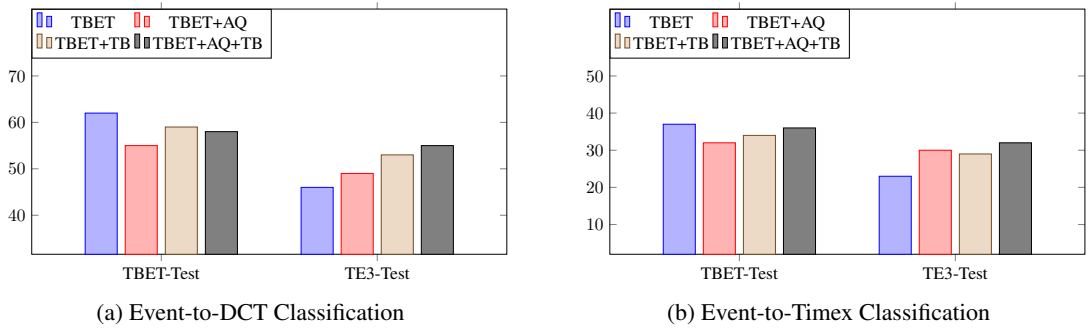
\begin{figure*}[t]
\centering
\subcaptionbox{Event-to-DCT Classification}{
\begin{tikzpicture}[scale=0.6, transform shape]
\centering
\begin{axis}[
    ymin=46,
    ymax=70,
    enlargelimits=0.6,
	ybar=5pt,
	bar width=20pt,
    height=7cm,
    width=12cm,
    legend style={at={(0,1)},
      anchor=north west,legend columns=2},
    symbolic x coords={TBET-Test, TE3-Test},
    ytick={30,40,50,60,70},
    yticklabel={\pgfmathprintnumber\tick},
    xtick=data,
    nodes near coords align={vertical},
    ]
\addplot coordinates {(TBET-Test,62) (TE3-Test,46)};
\addplot coordinates {(TBET-Test,55) (TE3-Test,49)};
\addplot coordinates {(TBET-Test,59) (TE3-Test,53)};
\addplot coordinates {(TBET-Test,58) (TE3-Test,55)};
\legend{TBET,TBET+AQ,TBET+TB,TBET+AQ+TB}
\end{axis}
\end{tikzpicture}}
\quad\quad
\subcaptionbox{Event-to-Timex Classification}{
\begin{tikzpicture}[scale=0.6, transform shape]
\centering
\begin{axis}[
    ymin=20,
    ymax=50,
    enlargelimits=0.6,
	ybar=5pt,
	bar width=20pt,
    height=7cm,
    width=12cm,
    legend style={at={(0,1)},
      anchor=north west,legend columns=2},
    symbolic x coords={TBET-Test, TE3-Test},
    ytick={10,20,30,40,50},
    yticklabel={\pgfmathprintnumber\tick},
    xtick=data,
    nodes near coords align={vertical},
    ]
\addplot coordinates {(TBET-Test,37) (TE3-Test,23)};
\addplot coordinates {(TBET-Test,32) (TE3-Test,30)};
\addplot coordinates {(TBET-Test,34) (TE3-Test,29)};
\addplot coordinates {(TBET-Test,36) (TE3-Test,32)};
\legend{TBET,TBET+AQ,TBET+TB,TBET+AQ+TB}
\end{axis}
\end{tikzpicture}}

\caption{The classification performance as the training data size increases. The evaluation metric is F1 of 4-SR 'complete match'.} \label{fig:largedata}
\end{figure*}

\subsection{Inference Strategy}

Once a target event given the predicted 4-SR to a Timex, the SRs naturally indicate the action of operating the normalized value of the Timex to infer the time of the event. For instance, if $SR_1$ is {\it equal} to a Timex, our inference strategy will update $(begin_e^e, begin_l^e)$ as the same as $(begin_e^t, begin_l^t)$. If $SR_1$ is {\it after}, the strategy will update $begin_e^e$ to the same as $begin_l^t$.

We infer the time of the target event along the SRs to DCT, Timex-0 (the same sentence), Timex-1 (the adjacent sentences), etc. An exception is that if an Event-to-DCT $SR_i$ is predicted as {\it equal}, the following $SR_i$ of Event-to-Timex will not effect anymore. The reason is that Event-to-DCT usually provides a majority of time information to predict event time in this task. We give it higher priority than Event-to-Timex to avoid wrong prediction of Event-to-Timex.

\section{A Large Event Time Corpus (LETC)}
\label{sec:corpus}

\begin{table}[!t]
  \begin{center}
    \begin{tabular}{l|c|c|c} 
      \textbf{Corpus} & \textbf{Article} & \textbf{Event} & \textbf{IAA}\\
      \hline
      TBET & 36 & 1.5K & 0.580 \\
      AQ & 73 & 4.4K & 0.567 \\
      TB & 147 & 5.2K & 0.552 \\
    \end{tabular}
    \caption{\label{tab:agree} The corpus statistics of our annotation data. 'TB' stands for 147 articles of the original TimeBank excluding 36 TBET articles.}
  \end{center}
\end{table}

TempEval-2013 released a cleaned TBAQ (TB:183 articles, AQ: 73 articles) for training and evaluation data TE3-Test with 20 articles. The existing TBET corpus is relatively small, which is based on the a sub-set of TB and contains 1.5K events in total.

As we discussed in Section~\ref{subsec:bg:eta}, the event time annotation has the advantage of linear annotation effort. The TempEval-2013 data are annotated according to the TimeML specification, which provides required information including annotated events, time expressions and normalized time values. We first followed the event time schema to specify an annotation guideline and re-annotated of the TBET corpus. Then, we revised our guideline to annotate the time anchors of all the events in TB (147 articles, excluding 36 TBET articles), AQ and TE3-Test. Our large event time corpus and annotation guideline will be released soon.\footnote{\url{https://github.com/rcfsd/LETC}}

Table~\ref{tab:agree} shows the Inter-Annotator-Agreements (IAAs) of our annotation of TBET, TB and AQ. We use Krippendorff’s $\alpha$ with the nominal metric to compute the complete match. Both TB and AQ shows the similiar $\alpha$ scores around $0.55$ to $0.56$, which is slightly lower than $\alpha = 0.58$ in TBET.

\section{Experiments}

\subsection{The Effect of Larger Training Data}
\label{subsec:dataincrease}

\begin{table*}[!t]
  \begin{center}
    \begin{tabular}{l|c|c} 
      \textbf{Model} & \textbf{Links} & \textbf{Event Time Acc.} \\
      \hline
      baseline & E-D & 17.4\% \\
      baseline & E-T & 7.2\% \\
      baseline & E-D + E-T & 20.8\% \\
      \hline
      SR attention classifier & E-D & 21.2\% \\
      SR attention classifier & E-T & 8.9\% \\
      SR attention classifier (Proposed) & E-D + E-T & 24.6\% \\
    \end{tabular}
    \caption{\label{tab:main} The performance of our models trained with LETC, evaluated on TE-Test.}
  \end{center}
\end{table*}

The TempEval-2013\footnote{\url{https://www.cs.york.ac.uk/semeval-2013/task1/index.php\%3Fid=data.html}} data consists of two training corpora (TB, AQ) and one test data TE3-Test. As TBET is based on a subset (36 articles) of TimeBank (183 articles), 'TB' stands for the other 147 articles in TB excluding TBET. We treat TBET as the baseline and gradually add AQ, TB and AQ + TB into the train data. The increase of the training data is expected to effectively improve the performance of the SR classifiers. We separately evaluate the classification results on TBET-Test (9 articles), TE3-Test (20 articles). In the experiments on TBET-Test, we use the same data split as the previous work\footnote{\url{https://github.com/nchambers/caevo/blob/master/src/main/java/caevo/Evaluate.java}}. In the experiments on TE3-Test, we randomly split 20\% training data as the validation set. The validation set is used to optimize the hyper-parameters and perform early stopping. 

Figure~\ref{fig:largedata} shows the classification performance against different training data size. We find TBET performs surprisingly well on TBET-Test. The increase of other corpora even reduces the classification performance. It's possibly because the 36 documents of TBET was manually selected with the close published dates and share some overlapped information between the train and test split. However, in the evaluation on the larger TE3-Test, TBET obtains the lowest scores. As the training data size increases, the classification performance is significantly improved. We believe TE3-Test can more correctly reflect the generalization ability of the classification, because the article topics are less dependent to the training data and test data size is larger. The overall classification scores in TBET-Test are objectively higher than TE3-Test. However, these number are not comparable due to the difference of the test data.

\begin{table}[!t]
  \begin{center}
    \begin{tabular}{l|c|r} 
      \textbf{Link Type} & \textbf{Acc.} & \textbf{Link}\\
      \hline
      \multicolumn{3}{l}{TE3\_Test}  \\ \hline
      E-D & 48.3\% & 1.0\\
      E-D + E-T (sw=0) & 63.1\% & 1.4\\
      E-D + E-T (sw=1) & 70.9\% & 1.9\\
      E-D + E-T (sw=2) & 73.5\% & 2.5\\ \hline
      \multicolumn{3}{l}{TBET\_Test}  \\ \hline
      E-D & 64.6\% & 1.0\\
      E-D + E-T (sw=0) & 70.4\% & 1.4\\
      E-D + E-T (sw=1) & 77.8\% & 2.1\\
      E-D + E-T (sw=2) & 78.4\% & 2.7\\ \hline
    \end{tabular}
    \caption{\label{tab:oracle} The oracle test of different Event-to-Timex sentence window settings. 'sw' denotes the sentence distance between an event and a Timex. 'Link' column denotes the average number of links (E-D and E-T) for each event.} 
  \end{center}
\end{table}

\subsection{Oracle Test with Gold SRs}

SRs between an event and a Timex are computed based the two qradruple vectors. This method has the capability of inducing SRs between an event to a Timex with long distance, which means our system can collect the Event-to-Timex information as much as we want.  However, in reality, the overall performance will be a trade-off between the amount of Event-to-Timex information and the classification accuracy. More SRs will bring more errors produced by the Event-to-Timex classifier. 

We perform the oracle tests in two test data, which let our system accept gold SRs with different distance settings between events to Timex. It helps us to select a reasonable sentence window. Table~\ref{tab:oracle} shows that the Event-to-Timex information in the adjacent sentences can achieve 70.98\% (TE3\_Test) and 77.81\% (TBET\_Test) upper-bound accuracy and larger sentence window will significantly increase the number of E-T links. Our SR classifiers are set to provide the information of Event-to-DCT and Event-to-Timex in two adjacent sentences for the inference strategy. 

Table~\ref{tab:oracle} also reveals that Event-to-DCT in TE3\_Test contributes approximate 16\% less than TBET\_Test to the event time prediction. Considering the relatively lower classification scores of TE3\_Test in Figure~\ref{fig:largedata}, we can expect that the event time prediction scores of TE3\_Test will certainly lower than TBET\_Test a lot. And of course, the classification and event time prediction performance of two test data are not comparable.

\subsection{Event Time Prediction in LETC}
\label{subsec:main}

\begin{table}[t]
  \begin{center}
    \begin{tabular}{l|c|c|c|c|c} 
      \textbf{Link} & $\bf SR_1$ & $\bf SR_2$ & $\bf SR_3$ & $ \bf SR_4$ & \textbf{Comp.} \\
      \hline
      E-D & 71.0 & 63.2 & 71.2 & 63.2 & 32.4 \\
      E-T & 52.7 & 61.5 & 59.6 & 58.7 & 55.2 \\
    \end{tabular}
    \caption{\label{tab:srres} The detailed classification F1scores of our proposed model trained on LETC. }
  \end{center}
\end{table}

We perform the experiment of training the SR classifiers with the LETC (TBET + AQ + TB) corpus and evaluate the performance in TE3-Test.

We first report the classification peformance of each SR and 'complete match' in Table~\ref{tab:srres}. The Event-to-DCT results show very similar scores between $SR_1$ and $SR_3$, $SR_2$ and $SR_4$. This is because DCT is always a \emph{Single-Day} and \emph{Certain} time anchor. It's quadruple vector has the same begin and end points. 

In Table~\ref{tab:main}, we report the performance of predicting event time. We separately feed the predicted SRs (E-D, E-T) into the inference strategy. The last column shows the 'complete match' accuracy of event time prediction. We provide a baseline model for observing the effect of the attention, which feeds the word and position embeddings into a Bidirectional LSTM model and feed the last hidden state into the Softmax layer. 

Event-to-DCT contributes to the main performance for predicting event time, which is matching the observation from the oracle test (Table~\ref{tab:oracle}). Our mention-wised attention model significantly outperforms the baseline, which used the same word and position embedding inputs.

\subsection{Comparing the SR classifier to SOTA}
\label{subsec:sota}

\begin{table}[!t]
  \vspace{15pt}
  \begin{center}
    \begin{tabular}{l|c} 
      \textbf{Model} & \textbf{Acc.} \\
      \hline
      Decision Tree~\cite{reimers2018event} & 41.6\% \\
      SR Classifier (Proposed) & 43.2\% \\
      SR Classifier w/o attn & 40.9\% \\
    \end{tabular}
    \caption{\label{tab:sota} The comparison to the SOTA model trained on TBET.}
  \end{center}
\end{table}

We compare our model to the SOTA decision tree model trained on the re-annotated TBET. The experiment follows the exactly same data split as the previous work: 22/5/9 articles as the train/dev/test data. Table~\ref{tab:sota} shows that our mention-wise attention classifier outperforms the decision tree model trained on the same data. The performance of the w/o attention model slightly lower than the decision tree model. 

Our system is relatively lightweight compared to the decision tree model, in which each decision node relies on a classifier, while our system consists of only two classifiers: Event-to-DCT and Event-to-Timex. 


\section{Conclusion}

In this paper, we proposed a simple but effective approach to address the event time prediction task. The existing model was hampered by the discrete time anchor representations and \emph{Uncertain} information, which increased the complexity of requiring the model to make hierarchical decisions. We first proposed a unified quadruple representation of time anchors. For preventing explicitly extending new temporal relation types to include the \emph{Uncertain} information, we proposed an idea of decomposing a temporal relation into four sub-level relations (SRs). Any complex relation can be derived as a combination of SRs. In the second step, we proposed a simple event time prediction system, which is composed of two mention-wised attention LSTM classifiers and an inference strategy. Our third contribution is the construction of a larger event time corpus of 256 news articles, compared to the existing 36 TBET corpus. The empirical results showed our proposed method outperformed the decision tree model with a more lightweight structure. Other experiments show that as the increase of training data size, the classification performance is significantly improved in a public test data: TE3-Test. 

This work leaves the space of improvement from several aspects. For instance, the current SRs adopt a coarse definition as the relation between the begin/end points between two time anchors. A fine-grained sub-level relation could be defined as a relation between one of the quadruple values, which will raise the number of SRs to 16 ($4 \times 4$). It can possibly encode more accurate temporal information.   Also this work follows the traditional temporal relation classification setting, which trained the Event-to-DCT and Event-to-Timex classifiers independently. Intuitively, Event-to-DCT and Event-to-Timex share the common useful information to be trained jointly or be optimized globally based on the event time prediction score. 

\bibliography{emnlp2020}

\begin{thebibliography}{21}
\expandafter\ifx\csname natexlab\endcsname\relax\def\natexlab#1{#1}\fi

\bibitem[{Allen(1990)}]{allen1990maintaining}
James~F Allen. 1990.
\newblock Maintaining knowledge about temporal intervals.
\newblock In \emph{Readings in qualitative reasoning about physical systems},
  pages 361--372. Elsevier.

\bibitem[{Berberich et~al.(2010)Berberich, Bedathur, Alonso, and
  Weikum}]{berberich2010language}
Klaus Berberich, Srikanta Bedathur, Omar Alonso, and Gerhard Weikum. 2010.
\newblock A language modeling approach for temporal information needs.
\newblock In \emph{European Conference on Information Retrieval}, pages 13--25.
  Springer.

\bibitem[{Cassidy et~al.(2014)Cassidy, McDowell, Chambers, and
  Bethard}]{cassidy2014}
Taylor Cassidy, Bill McDowell, Nathanael Chambers, and Steven Bethard. 2014.
\newblock \href {http://www.aclweb.org/anthology/P14-2082} {An annotation
  framework for dense event ordering}.
\newblock In \emph{Proceedings of the 52nd Annual Meeting of the Association
  for Computational Linguistics (Volume 2: Short Papers)}, pages 501--506,
  Baltimore, Maryland. Association for Computational Linguistics.

\bibitem[{Chambers et~al.(2014)Chambers, Cassidy, McDowell, and
  Bethard}]{chambers2014dense}
Nathanael Chambers, Taylor Cassidy, Bill McDowell, and Steven Bethard. 2014.
\newblock \href {http://aclweb.org/anthology/Q/Q14/Q14-1022.pdf} {Dense event
  ordering with a multi-pass architecture}.
\newblock \emph{Transactions of the Association for Computational Linguistics},
  2:273--284.

\bibitem[{Cheng and Miyao(2017)}]{cheng-miyao:2017:Short}
Fei Cheng and Yusuke Miyao. 2017.
\newblock \href {http://aclweb.org/anthology/P17-2001} {Classifying temporal
  relations by bidirectional lstm over dependency paths}.
\newblock In \emph{Proceedings of the 55th Annual Meeting of the Association
  for Computational Linguistics (Volume 2: Short Papers)}, pages 1--6,
  Vancouver, Canada. Association for Computational Linguistics.

\bibitem[{Cheng and Miyao(2018)}]{N18-1166}
Fei Cheng and Yusuke Miyao. 2018.
\newblock \href {https://doi.org/10.18653/v1/N18-1166} {Inducing temporal
  relations from time anchor annotation}.
\newblock In \emph{Proceedings of the 2018 Conference of the North American
  Chapter of the Association for Computational Linguistics: Human Language
  Technologies, Volume 1 (Long Papers)}, pages 1833--1843. Association for
  Computational Linguistics.

\bibitem[{Hochreiter and Schmidhuber(1997)}]{hochreiter1997long}
Sepp Hochreiter and J{\"u}rgen Schmidhuber. 1997.
\newblock \href {https://doi.org/10.1162/neco.1997.9.8.1735} {Long short-term
  memory}.
\newblock \emph{Neural computation}, 9(8):1735--1780.

\bibitem[{Llorens et~al.(2015)Llorens, Chambers, UzZaman, Mostafazadeh, Allen,
  and Pustejovsky}]{llorens-EtAl:2015:SemEval}
Hector Llorens, Nathanael Chambers, Naushad UzZaman, Nasrin Mostafazadeh, James
  Allen, and James Pustejovsky. 2015.
\newblock \href {http://www.aclweb.org/anthology/S15-2134} {Semeval-2015 task
  5: Qa tempeval - evaluating temporal information understanding with question
  answering}.
\newblock In \emph{Proceedings of the 9th International Workshop on Semantic
  Evaluation (SemEval 2015)}, pages 792--800, Denver, Colorado. Association for
  Computational Linguistics.

\bibitem[{Luong et~al.(2015)Luong, Pham, and
  Manning}]{luong-etal-2015-effective}
Thang Luong, Hieu Pham, and Christopher~D. Manning. 2015.
\newblock \href {https://doi.org/10.18653/v1/D15-1166} {Effective approaches to
  attention-based neural machine translation}.
\newblock In \emph{Proceedings of the 2015 Conference on Empirical Methods in
  Natural Language Processing}, pages 1412--1421, Lisbon, Portugal. Association
  for Computational Linguistics.

\bibitem[{Meng et~al.(2017)Meng, Rumshisky, and
  Romanov}]{meng-rumshisky-romanov:2017:EMNLP2017}
Yuanliang Meng, Anna Rumshisky, and Alexey Romanov. 2017.
\newblock \href {https://www.aclweb.org/anthology/D17-1092} {Temporal
  information extraction for question answering using syntactic dependencies in
  an lstm-based architecture}.
\newblock In \emph{Proceedings of the 2017 Conference on Empirical Methods in
  Natural Language Processing}, pages 887--896, Copenhagen, Denmark.
  Association for Computational Linguistics.

\bibitem[{Minard et~al.(2015)Minard, Speranza, Agirre, Aldabe, van Erp,
  Magnini, Rigau, and Urizar}]{minard-EtAl:2015:SemEval}
Anne-Lyse Minard, Manuela Speranza, Eneko Agirre, Itziar Aldabe, Marieke van
  Erp, Bernardo Magnini, German Rigau, and Ruben Urizar. 2015.
\newblock \href {http://www.aclweb.org/anthology/S15-2132} {Semeval-2015 task
  4: Timeline: Cross-document event ordering}.
\newblock In \emph{Proceedings of the 9th International Workshop on Semantic
  Evaluation (SemEval 2015)}, pages 778--786, Denver, Colorado. Association for
  Computational Linguistics.

\bibitem[{Pennington et~al.(2014)Pennington, Socher, and
  Manning}]{pennington2014glove}
Jeffrey Pennington, Richard Socher, and Christopher~D. Manning. 2014.
\newblock \href {http://www.aclweb.org/anthology/D14-1162} {Glove: Global
  vectors for word representation}.
\newblock In \emph{Empirical Methods in Natural Language Processing (EMNLP)},
  pages 1532--1543.

\bibitem[{Pustejovsky et~al.(2003)Pustejovsky, Hanks, Sauri, See, Gaizauskas,
  Setzer, Radev, Sundheim, Day, Ferro et~al.}]{pustejovsky2003timebank}
James Pustejovsky, Patrick Hanks, Roser Sauri, Andrew See, Robert Gaizauskas,
  Andrea Setzer, Dragomir Radev, Beth Sundheim, David Day, Lisa Ferro, et~al.
  2003.
\newblock \href {https://catalog.ldc.upenn.edu/LDC2006T08} {The timebank
  corpus}.
\newblock In \emph{Corpus linguistics}, volume 2003, page~40.

\bibitem[{Reimers et~al.(2016)Reimers, Dehghani, and
  Gurevych}]{reimers-dehghani-gurevych:2016:P16-1}
Nils Reimers, Nazanin Dehghani, and Iryna Gurevych. 2016.
\newblock \href {http://www.aclweb.org/anthology/P16-1207} {Temporal anchoring
  of events for the timebank corpus}.
\newblock In \emph{Proceedings of the 54th Annual Meeting of the Association
  for Computational Linguistics (Volume 1: Long Papers)}, pages 2195--2204,
  Berlin, Germany. Association for Computational Linguistics.

\bibitem[{Reimers et~al.(2018)Reimers, Dehghani, and
  Gurevych}]{reimers2018event}
Nils Reimers, Nazanin Dehghani, and Iryna Gurevych. 2018.
\newblock Event time extraction with a decision tree of neural classifiers.
\newblock \emph{Transactions of the Association of Computational Linguistics},
  6:77--89.

\bibitem[{Setzer(2002)}]{setzer2002temporal}
Andrea Setzer. 2002.
\newblock \emph{Temporal information in newswire articles: an annotation scheme
  and corpus study.}
\newblock Ph.D. thesis, University of Sheffield.

\bibitem[{UzZaman et~al.(2012)UzZaman, Llorens, Allen, Derczynski, Verhagen,
  and Pustejovsky}]{uzzaman2012tempeval}
Naushad UzZaman, Hector Llorens, James Allen, Leon Derczynski, Marc Verhagen,
  and James Pustejovsky. 2012.
\newblock \href {http://aclweb.org/anthology/S/S13/S13-2001.pdf} {Tempeval-3:
  Evaluating events, time expressions, and temporal relations}.
\newblock \emph{arXiv preprint arXiv:1206.5333}.

\bibitem[{Verhagen et~al.(2009)Verhagen, Gaizauskas, Schilder, Hepple,
  Moszkowicz, and Pustejovsky}]{verhagen2009tempeval}
Marc Verhagen, Robert Gaizauskas, Frank Schilder, Mark Hepple, Jessica
  Moszkowicz, and James Pustejovsky. 2009.
\newblock \href {https://doi.org/10.1007/s10579-009-9086-z} {The tempeval
  challenge: identifying temporal relations in text}.
\newblock \emph{Language Resources and Evaluation}, 43(2):161--179.

\bibitem[{Verhagen et~al.(2010)Verhagen, Sauri, Caselli, and
  Pustejovsky}]{verhagen2010semeval}
Marc Verhagen, Roser Sauri, Tommaso Caselli, and James Pustejovsky. 2010.
\newblock \href {http://www.aclweb.org/anthology/S10-1010} {Semeval-2010 task
  13: Tempeval-2}.
\newblock In \emph{Proceedings of the 5th international workshop on semantic
  evaluation}, pages 57--62. Association for Computational Linguistics.

\bibitem[{Zeng et~al.(2014)Zeng, Liu, Lai, Zhou, and Zhao}]{C14-1220}
Daojian Zeng, Kang Liu, Siwei Lai, Guangyou Zhou, and Jun Zhao. 2014.
\newblock \href {http://aclweb.org/anthology/C14-1220} {Relation classification
  via convolutional deep neural network}.
\newblock In \emph{Proceedings of COLING 2014, the 25th International
  Conference on Computational Linguistics: Technical Papers}, pages 2335--2344.
  Dublin City University and Association for Computational Linguistics.

\bibitem[{Zhou et~al.(2016)Zhou, Shi, Tian, Qi, Li, Hao, and
  Xu}]{zhou-etal-2016-attention}
Peng Zhou, Wei Shi, Jun Tian, Zhenyu Qi, Bingchen Li, Hongwei Hao, and Bo~Xu.
  2016.
\newblock \href {https://doi.org/10.18653/v1/P16-2034} {Attention-based
  bidirectional long short-term memory networks for relation classification}.
\newblock In \emph{Proceedings of the 54th Annual Meeting of the Association
  for Computational Linguistics (Volume 2: Short Papers)}, pages 207--212,
  Berlin, Germany. Association for Computational Linguistics.

\end{thebibliography}
\bibliographystyle{acl_natbib}

\end{document}